\documentclass[a4paper,twoside]{article}

\usepackage{epsfig}
\usepackage{subcaption}
\usepackage{calc}
\usepackage{amssymb}
\usepackage{amstext}
\usepackage{amsmath}
\usepackage{amsthm}
\usepackage{multicol}
\usepackage{pslatex}
\usepackage{apalike}
\usepackage{algorithm2e}
\usepackage{booktabs}
\usepackage[bottom]{footmisc}
\usepackage{SCITEPRESS}     
\usepackage[outdir=./]{epstopdf}

\newcommand{\etal}{{\em et al. }}

\begin{document}

\title{Attacking the Loop: Adversarial Attacks on \\Graph-based Loop Closure Detection}

\author{\authorname{Jonathan J.Y. Kim\sup{1,3}\orcidAuthor{0000-0003-4287-4884}, Martin Urschler\sup{1,2}\orcidAuthor{0000-0001-5792-3971}, Patricia J. Riddle\sup{1}\orcidAuthor{0000-0001-8616-0053} and J\"org S. Wicker\sup{1}\orcidAuthor{0000-0003-0533-3368}}
\affiliation{\sup{1}School of Computer Science, University of Auckland, New Zealand}
\affiliation{\sup{2}Institute for Medical Informatics, Statistics and Documentation, Medical University Graz, Austria}
\affiliation{\sup{3}Callaghan Innovation, Auckland, New Zealand}
\email{jkim072@aucklanduni.ac.nz, martin.urschler@medunigraz.at, \{p.riddle,j.wicker\}@auckland.ac.nz}
}



\abstract{With the advancement in robotics, it is becoming increasingly common for large factories and warehouses to incorporate visual SLAM (vSLAM) enabled automated robots that operate closely next to humans. This makes any adversarial attacks on vSLAM components potentially detrimental to humans working alongside them. 
Loop Closure Detection (LCD) is a crucial component in vSLAM that minimizes the accumulation of drift in mapping, since even a small drift can accumulate into a significant drift over time.
A prior work by Kim \textit{et al.}, SymbioLCD2, unified visual features and semantic objects into a single graph structure for finding loop closure candidates. While this provided a performance improvement over visual feature-based LCD, it also created a single point of vulnerability for potential graph-based adversarial attacks. Unlike previously reported visual-patch based attacks, small graph perturbations are far more challenging to detect, making them a more significant threat. 
In this paper, we present \textit{Adversarial-LCD}, a novel \textit{black-box} \textit{evasion} attack framework that employs an eigencentrality-based perturbation method and an SVM-RBF surrogate model with a Weisfeiler-Lehman feature extractor for attacking graph-based LCD. 
Our evaluation shows that the attack performance of \textit{Adversarial-LCD} with the SVM-RBF surrogate model was superior to that of other machine learning surrogate algorithms, including SVM-linear, SVM-polynomial, and Bayesian classifier, demonstrating the effectiveness of our attack framework. 
Furthermore, we show that our eigencentrality-based perturbation method outperforms other algorithms, such as Random-walk and Shortest-path, highlighting the efficiency of \textit{Adversarial-LCD}'s perturbation selection method.}

\onecolumn \maketitle \normalsize \setcounter{footnote}{0} \vfill

\section{\uppercase{Introduction}}
\label{sec:introduction}

Simultaneous Localization and Mapping (SLAM) refers to a technique that involves creating a map of an unknown environment while simultaneously determining a device's location within the map.
Visual SLAM (vSLAM) refers to a subset of SLAM being performed only using visual sensors \cite{ORBSLAM,dynaslam,reslam}. 
With recent advancements in robotics, it is becoming increasingly common for large factories and warehouses to incorporate vSLAM-enabled automated robots into their operations. These robots are often designed to operate in close proximity with human workers. 

\begin{figure}[t]
\begin{center}
\includegraphics[width=1.0\linewidth]{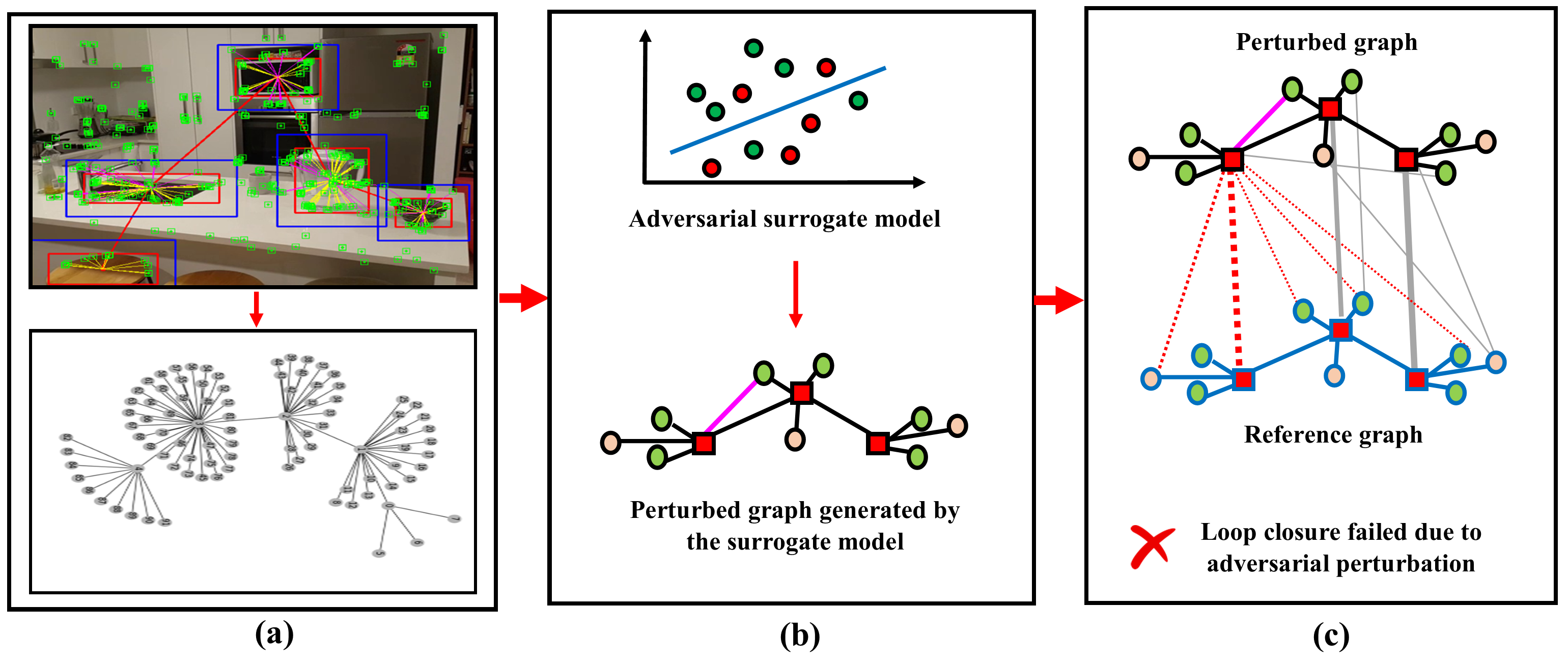}
\end{center}
\caption{Basic overview of Adversarial-LCD (a) an input image is transformed into a unified graph structure (b) adversarial attacks are performed using a surrogate model (c) the target model is adversely affected by adversarial attacks.}
\label{fig:title}
\end{figure}

As a result, it is becoming essential to thoroughly examine vSLAM frameworks for potential vulnerabilities using various methods, such as adversarial attacks, to ensure the safety of human operators in such collaborative settings \cite{Ikram}.

Adversarial attacks \cite{adv_struct,Wan} refer to a class of attacks that are designed to exploit vulnerabilities in Machine Learning (ML) models or systems by intentionally manipulating input data in a way that causes the model or system to produce incorrect results. 
A \textit{white-box} attack \cite{white} is a type of adversarial attack that is carried out with complete knowledge of the target models' internal parameters and training data. 
In a \textit{black-box} attack \cite{atk_ratio}, the attacker has no knowledge of the target models' internal parameters or training data. They can only interact with the target model by querying input data and observing its output. An \textit{evasion} attack \cite{Wan} refers to the technique of creating an adversarial example by adding imperceptible perturbations to input data to adversely affect the target model.

Ikram \etal \cite{Ikram} have shown that adversarial attacks can impact Loop Closure Detection (LCD) in vSLAM systems. They demonstrated this by modifying the environment, \textit{i.e.} placing a simple high-textured patch in different places in the physical scene, thus adversely affecting the visual feature matching in LCD.
However, since the attack requires putting visual patches in multiple places, it has the potential to be discovered by human workers nearby.

The key challenge of vSLAM is to estimate the device trajectory accurately, even in the presence of feature matching inconsistencies and of \textit{drift} from a sensor, as even small amounts of drift can accumulate to become potentially substantial by the end of the trajectory.
To overcome the \textit{drift} issue, vSLAM uses loop closure detection. It is a process of recognizing and correcting drift in the trajectory by revisiting previously visited locations, and it is essential for keeping consistent location and mapping within vSLAM.

SymbioLCD2 \cite{kim2022closing} simplified an image's semantic and spatial associations by developing a unified graph structure that integrates visual features, semantic objects, and their spatial relationship. 
While the development of a unified graph structure has improved performance over visual feature-based LCD \cite{ORBSLAM,dynaslam}, it has also introduced a potential point of vulnerability, where a malicious actor could carry out graph-based adversarial attacks to affect the LCD process adversely.
Unlike visual patch-based attacks, graph-based attacks pose a higher threat, as small perturbations in a graph would be much harder to detect unless someone closely examines each graph instance.

\begin{figure*}[ht]
\begin{center}
\includegraphics[width=1.0\linewidth]{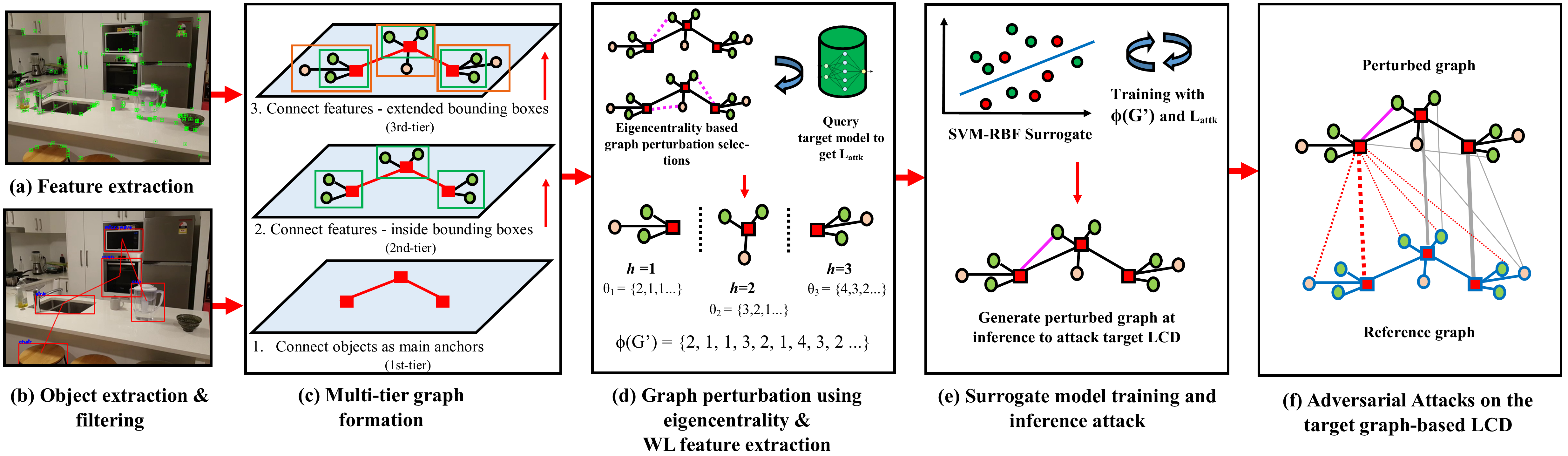}
\end{center}
   \caption{Detailed overview of \textit{Adversarial-LCD}. (a) Visual feature extraction and generating a vBoW score. (b) Semantic object extraction from CNN and object filtering based on their location and size. (c) Multi-tier graph formation using semantic objects as main anchors - object and feature information gets transferred as node features, and distances between objects and features become edge features. (d) Graph perturbation selection using eigencentrality. The features from perturbed graphs are extracted using the WL feature extractor. (e) Training the surrogate model and generating perturbed graphs for adversarial attacks on the target model. The purple edge indicates an example perturbation. (f) Adversarial attacks are performed on the target graph-based LCD, causing it to perform incorrect loop closure.}
\label{fig:det_overview}
\end{figure*}

In this paper, we study graph-based attacks by proposing our novel \textit{black-box} \textit{evasion} attack framework, called \textit{Adversarial-LCD}, which employs an eigencentrality graph perturbation method, a Support Vector Machine (SVM) with Radial Basis Function (RBF) Kernel surrogate model, and a Weisfeiler-Lehman (WL) feature extractor. 
Firstly, it utilizes an eigencentrality perturbation method to select graph perturbations efficiently. This is accomplished by identifying the most well-connected nodes, which correspond to the most influential connections.
Secondly, the WL feature extractor generates concatenated feature vectors from the perturbed graphs. This enables the surrogate model to learn directly from the graph-like search space.
Thirdly, the framework incorporates an SVM-RBF surrogate model, which offers highly efficient performance even with a small number of training datasets. 
Figure \ref{fig:title} shows the basic framework and Figure \ref{fig:det_overview} shows a detailed overview of our proposed \textit{Adversarial-LCD}.

The main contributions of this paper are as follows:
\begin{itemize}

\item To the best of our knowledge, this is the first work presenting adversarial attacks on graph-based Loop Closure Detection. 
\item We propose \textit{Adversarial-LCD}, a \textit{black-box} \textit{evasion} attack framework using an eigencentrality graph perturbation method and an SVM-RBF surrogate model with a WL feature extractor.
\item We show that our \textit{Adversarial-LCD} with the SVM-RBF surrogate model outperforms other ML surrogate algorithms, such as SVM-linear, SVM-polynomial and Bayesian classifier.
\item We show that our \textit{Adversarial-LCD} with eigencentrality graph perturbation method is more efficient than other perturbation methods, such as random-walk and shortest-path.

\end{itemize}

This paper is organized as follows. Section \ref{sec:relatedwork} reviews related work, Section \ref{sec:proposedmethod} demonstrates our proposed method, Section \ref{sec:experiments} shows experiments and their results, and Section \ref{sec:conclusion} concludes the paper.

\section{\uppercase{Related Work}}
\label{sec:relatedwork}
This section reviews graph neural networks, graph perturbation methods, adversarial attacks on graph neural networks and adversarial attacks on loop closure detection.

\textbf{Graph Neural Networks}
Graph-structured data is a useful way of representing spatial relationships in a scene. Graph Neural Networks (GNNs) have become increasingly popular for efficiently learning relational representations in graph-structured data \cite{graphsage}. GNN models are widely used for graph classification \cite{ete} and can generate graph embeddings in vector spaces to predict the similarity between a pair of graphs, making similarity reasoning more efficient \cite{GMN,GAN}.

In addition to GNNs, graph kernels \cite{grakel} have emerged as a promising approach for graph classification. They enable kernelized learning algorithms, such as SVM and WL, to perform attributed subgraph matching and achieve state-of-the-art performance on graph classification tasks \cite{grakel}. The WL graph kernel utilizes a unique labelling scheme to extract subgraph patterns through multiple iterations. Each node's label is replaced with a label that consists of its original label and the subset of labels of its direct neighbours, which is then compressed to form a new label. The similarity between the two graphs is calculated as the inner product of their histogram vectors after the relabeling procedures \cite{weis}.

\textbf{Graph perturbation methods}
There are several graph perturbation methods that have been proposed for adversarial attacks on graph-based models \cite{adv_struct,Wan}. A graph perturbation aims to create a new graph that is similar to the original graph but with slight changes that can deceive the model into making incorrect predictions.

Random edge perturbation \cite{Wan} involves randomly adding or removing edges from the graph to perturb the relationships between nodes. These perturbations in the graph increase the likelihood of false prediction on the target model. However, since the changes are made randomly, there is no guarantee that they will be effective in deceiving the model.

Shortest path perturbation \cite{short2} involves modifying the shortest path between two nodes in the graph by adding or removing edges to create a longer or shorter path. This can cause the model to make incorrect predictions by changing the relationships between nodes in the graph. This method is more targeted than random edge perturbation and has shown to be more effective in some cases \cite{short2}.

Eigencentrality perturbation \cite{ec} involves modifying the centrality of nodes in the graph based on their eigencentrality, which is a measure of their importance in the network. This method targets the most important nodes in the graph and can have a significant impact on the model's predictions. Our graph perturbation method is based on eigencentrality.

\textbf{Adversarial attacks on Graph Neural Networks}
In recent years, GNNs have gained significant attention and have been instrumental in various fields. However, like other deep neural networks, GNNs are also susceptible to malicious attacks. Adversarial attacks involve creating deceptive examples with minimal perturbations to the original data to reduce the performance of target models. 

A \textit{white-box} attack \cite{white} is an adversarial attack that is carried out with full knowledge of the internal parameters and training data of the target model. In a \textit{black-box} attack \cite{atk_ratio}, the attacker has no information about the target model's internal parameters or training data. They can only interact with the target model by querying input data and observing the model's output.

\textit{Poisoning} attacks \cite{adv_struct} involves deliberately injecting harmful samples during training to exploit the target machine learning model. Unlike data \textit{poisoning}, \textit{evasion} attacks \cite{evasion2} generate adversarial perturbations to input data to adversely affect the target model.

In a \textit{black-box} \textit{evasion} attack \cite{Wan} \cite{evasion2}, the attacker estimates the gradients of the target model's decision boundary with respect to the input data by repeatedly querying the target model with perturbed inputs and observing the corresponding outputs. Our method is a \textit{black-box} \textit{evasion} attack.

\textbf{Adversarial Attacks on Loop Closure Detection}
Adversarial attacks on LCD are still very nascent. 
Ikram \etal \cite{Ikram} investigated the impact of adversarial attacks on LCD in vSLAM systems. They demonstrated that modifying the environment by placing a simple high-textured patch in various locations can negatively affect visual feature matching in LCD. 
However, to the best of our knowledge, there is no adversarial attack performed on a graph-based LCD. Graph-based attacks pose a higher threat, as unlike visible patches, small perturbations in a graph would be much harder to detect unless someone closely examines each graph instance.

\section{\uppercase{Proposed Method}}
\label{sec:proposedmethod}
The overview of our proposed method is shown in Figure \ref{fig:det_overview}, and the reader may refer to \cite{kim2022closing} for further details on (a), (b) and (c).
Our proposed methods are as follows - firstly, the original input graph from Figure \ref{fig:det_overview} (c) is perturbed using the eigencentrality perturbation method. The target model is queried with the perturbed graphs, and observed attack losses are sent to the WL feature extraction process.
Secondly, the WL feature extraction process extracts concatenated feature vectors from the set of perturbed graphs. This approach enables the surrogate model to learn on the graph-like search space efficiently.
Lastly, the SVM-RBF surrogate model is trained using the extracted WL features of perturbed graphs and their corresponding attack losses. At inference time, we use the surrogate model to attack the target graph-based LCD to degrade its performance.
\subsection{Framework}
\label{sec:framework}
We propose \textit{Adversarial-LCD} as the framework for incorporating our adversarial attacks into the LCD framework. 
The \textit{Adversarial-LCD} framework was created using SymbioLCD2 \cite{kim2022closing}, which uses a unified graph structure as an input to its Weisfeiler-Lehman subgraph matching algorithm to predict loop closure candidates. 

The \textit{Adversarial-LCD} is made up of three modules. The first module, which we call the \textit{graph-module}, does the visual and semantic object extraction to create a multi-tier graph, as shown in Figure \ref{fig:det_overview}(a - c). The third module, which we call the \textit{target-LCD}, contains the loop closure detection model shown in Figure \ref{fig:det_overview}(f). 
The second module, which we call the \textit{attack-module}, shown in Figure \ref{fig:det_overview}(d \& e), is situated between the first and third modules. The \textit{attack-module} has access to the input graph $G$ as it is being sent from the \textit{graph-module} to the \textit{target-LCD}.  

\subsection{Problem Setup}
\label{sec:prob_setup}

We perform \textit{black-box} \textit{evasion} attacks, where our \textit{attack-module} has no access to the training data or parameters on the \textit{target-LCD} model $f_{t}$. However, it has access to the input graph $G$ from the \textit{graph-module}, and it can query the \textit{target-LCD} with a perturbed input graph $G'$ and observe the \textit{target-LCD} model output $f_t(G')$. 

Our adversarial attack aims to degrade the predictive performance of the \textit{target-LCD} via a \textit{black-box} maximization,
\begin{equation}
max_{G'} \ \mathcal{L}_{attk}(f_t(G'),y),
\end{equation}
where $\mathcal{L}_{attk}$ is the attack loss function, $G'$ is the perturbed version of $G$, and $y$ refers to the correct label of the original graph $G$.
\subsection{Perturbation selections using eigencentrality}
\label{sec:perturbation}
Changing a large number of graph connections can be expensive and runs the chance of being detected if the perturbation amount becomes excessive. Therefore, we use eigencentrality for effectively selecting the perturbations within the perturbation budget of $\beta = rn^2$, where $r$ refers to \textit{perturbation ratio} \cite{atk_ratio}, and $n$ refers to the number of nodes. 
Eigencentrality can identify the most well-connected nodes, i.e., the most influential connections, which have a higher chance of disruption when the connections are added or subtracted. 

Given the input graph $G=(V,E)$ with $v$ vertices and its neighbouring vertices $u$, an adjacency matrix $A$ could be defined as a set of $(a_v,u)$, where $a_v=1$ if the vertex $v$ is connected to the vertex $u$, or $a_v=0$ if it is not connected to the vertex $u$.
The eigencentrality score $X$ for a vertex $v$ can be defined as,
\begin{equation}
X_v = \frac{1}{\lambda} \sum_{u\epsilon V} A X_u,
\end{equation}

or in vector notation, 
\begin{equation}
Ax = \lambda x,
\end{equation}
where $\lambda$ is a constant. 
The eigencentrality amplifies the components of the vector corresponding to the largest eigenvalues, i.e. \textit{centrality}. 
We use the eigencentrality to iteratively generate a set of perturbed graphs and query the \textit{target-LCD} model $f_t(G')$ until the attack is successful, or until the maximum query budget is exhausted.
The resulting perturbed graphs, the original graph and their attack losses are sent to the Weisfeiler-Lehman feature extraction process.

\subsection{Weisfeiler-Lehman feature extraction}
\label{sec:WLextraction}

Building on the work of Wan \etal \cite{Wan}, we leverage the WL feature extractor to extract concatenated feature vectors from the set of perturbed graphs generated in the previous step. This approach enables us to directly train a surrogate model on the graph-like search space in an efficient manner. 
Given the initial node feature $x_0(v)$ of node $v$, WL feature extractor iteratively aggregates features of $v$ with features of its neighbour $u$, 
\begin{equation}
x^{h}_0(v) = aggregate(x^h(v),x^h(u_1),...,x^h(u_i)),
\end{equation}
where $h$ refers to the number of iterations. 
At each $h$, feature vector $\phi_h(G')$ can be defined as,
\begin{equation}
\phi_h(G') = (x^h_0(v),x^h_1(v),...,x^h_i(v)).
\end{equation}

The final feature vector at the end of the total number of iterations $H$,
\begin{equation}
\phi(G') = concat(\phi_1(G'),\phi_2(G'),...,\phi_H(G')),
\end{equation}
is sent to the surrogate model for training.

\subsection{Surrogate model}
\label{sec:surrogate}

The SVM is a widely recognized ML algorithm for its simplicity and effectiveness in finding the optimal hyperplane. It also offers kernel functions, which are a potent tool for navigating high-dimensional spaces. With kernel functions, SVM can directly map the data into higher dimensions without the need to transform the entire dataset \cite{gaussian}.
Thus, we utilize SVM-RBF as our surrogate model as it can deliver efficient training performance with Gaussian probabilistic output in a binary classification setting.
We train our SVM-RBF surrogate with WL feature vectors $\phi(G')$ and their attack losses $y'=\mathcal{L}_{attk}$ as inputs.

RBF combines various polynomial kernels with differing degrees to map the non-linear data into a higher dimensional space, so that it can be separated using a hyperplane.
RBF kernel maps the data into a higher-dimensional space by,
\begin{equation}
K(\phi(G'_i),\phi(G'_j))=\exp(-\frac{||\phi(G'_i)-\phi(G'_j)||^2}{2\sigma^2}),
\end{equation}
where $\sigma$ is a tuning parameter, based on the standard deviation of a dataset.
To simplify, we assume $\gamma=\frac{1}{2\sigma^2}$, which leads to,
\begin{equation}
K(\phi(G'_i),\phi(G'_j))=\exp(-\gamma||\phi(G'_i)-\phi(G'_j)||^2).
\end{equation}

With the kernel function, the optimization of the SVM surrogate model can be written as,
\begin{equation}
min_\alpha \frac{1}{2}\sum^N_{i=1}\sum^N_{j=1} \alpha_i \alpha_j y'_i y'_j K(\phi(G'_i),\phi(G'_j)) - \sum^N_{i=1} \alpha_i, 
\end{equation}
\begin{equation}
s.t. \sum^N_{i=1} \alpha_i y'_i = 0 \ \& \ \alpha_i \geq 0,
\end{equation}

where $\alpha$ refers to a Lagrange multiplier \cite{Lagrange} corresponding to the training data $\phi(G')$.

\section{\uppercase{Experiments}}
\label{sec:experiments}

We have evaluated \textit{Adversarial-LCD} with the following experiments.
Section \ref{sec:datasets} shows datasets and evaluation parameters used in the experiments.
Section \ref{sec:accuracy} evaluates \textit{Adversarial-LCD} with the SVM-RBF surrogate model against other machine learning surrogate models.
Section \ref{sec:eigen} evaluates the eigencentrality perturbation method against other perturbation algorithms. 

\subsection{Setup}
\label{sec:datasets}
\begin{figure*}[ht]
\begin{center}
\includegraphics[width=1.0\linewidth]{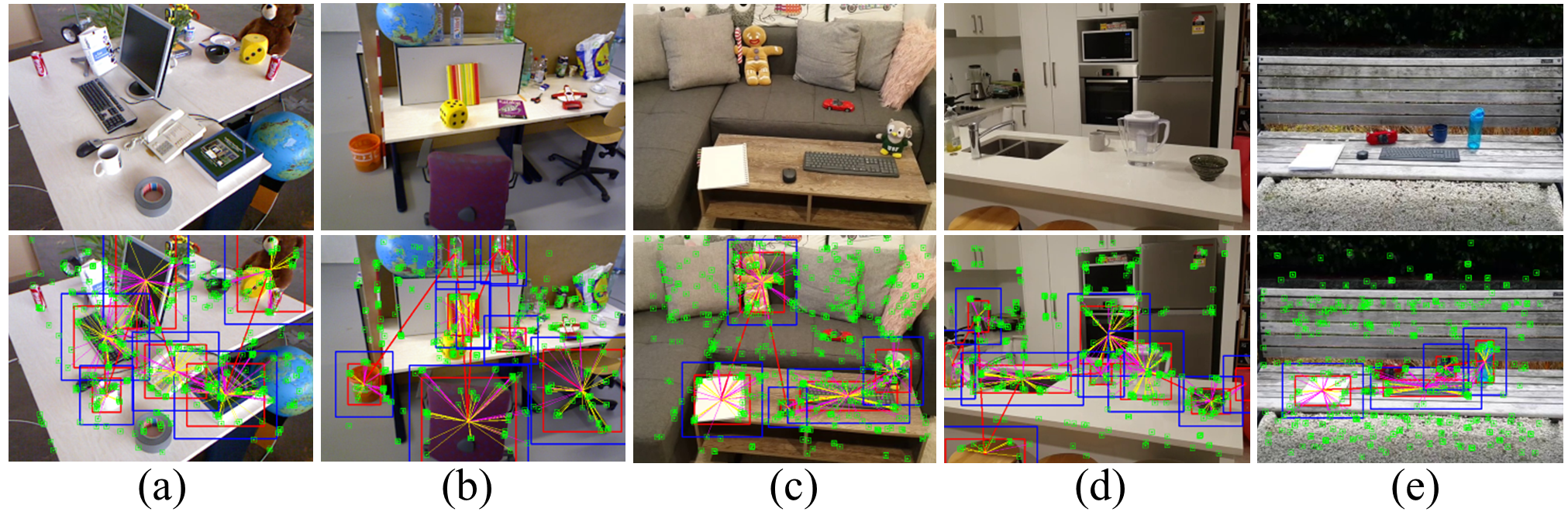}
\end{center}
  \caption{Evaluation datasets. (a) fr2-desk (b) fr3-longoffice (c) uoa-lounge (d) uoa-kitchen (e) uoa-garden}
  \label{fig:datasets}
\end{figure*}

For evaluating our \textit{Adversarial-LCD}, we have selected five publicly available datasets with multiple objects and varying camera trajectories. 
We selected fr2-desk and fr3-longoffice from the TUM dataset \cite{TUM_data}, and uoa-lounge, uoa-kitchen and uoa-garden from the University of Auckland multi-objects dataset \cite{symbio}. The details of the datasets are shown in Table \ref{tab:dataset}. Table \ref{tab:parameters} shows evaluation parameters and Figure \ref{fig:datasets} shows examples from each dataset. All experiments were performed on a PC with Intel i9-10885 and Nvidia GTX2080.

\begin{table} [ht]
\tiny
\caption{Parameters and Datasets}
 \begin{subtable}[h]{0.35\columnwidth}
 \tiny
 \scriptsize
 \centering
  \begin{tabular}{cc} \toprule \midrule
      Parameters & Value    \\ \midrule
      Epochs     & 200 \\
      Rand. state & 42 \\
      \textit{r} & $3e^{-4}$   \\
      $\lambda$    & 0.1  \\
      $\gamma$    & $\frac{1}{2\sigma^2}$  \\
      $\alpha$     &0.05  \\
     \midrule 
  \end{tabular}
  \caption{Parameters}
  \label{tab:parameters}
 \end{subtable}
 \hfill
 \begin{subtable}[h]{0.64\columnwidth}
 \tiny
 \centering
 \scriptsize
  \begin{tabular}{cccc} \toprule \midrule
      Dataset & Source & No. of  & Image   \\
      & & frames & Res.\\
      \midrule
      fr2-desk       & TUM  & 2965   &   640x480\\
      fr3-long. & TUM   & 2585  &   640x480\\
      lounge         & ours & 1841   &   640x480\\
      kitchen         & ours & 1998  &   640x480\\
      garden         & ours & 2148   &   640x480\\ \midrule 
  \end{tabular}
  \caption{Datasets descriptions}
  \label{tab:dataset}
 \end{subtable}
\end{table}

\subsection{Evaluation against other machine learning surrogate models}
\label{sec:accuracy}

We conducted a benchmark of \textit{Adversarial-LCD} with the SVM-RBF surrogate model against three other ML surrogate models, SVM-linear, SVM-polynomial, and Bayesian classifier. To evaluate each surrogate model, we attacked the \textit{target-LCD} model and recorded the decline in its accuracy. To simulate a realistic scenario where a large number of changes to the graph connections would easily raise suspicion, we allowed only a small perturbation budget of $r=3e^{-4}$ for the experiment. To account for the non-deterministic nature of the algorithms, we performed the evaluation ten times. The results, presented in Table \ref{tab:surrogate}, indicate that on average, \textit{Adversarial-LCD} with SVM-RBF achieved the highest decline in accuracy compared to the other algorithms, surpassing SVM-linear by 12.6\%, SVM-polynomial by 7.3\%, and Bayesian classifier by 2.7\%. 

\begin{figure}[h]
\begin{center}
\includegraphics[width=1.0\linewidth]{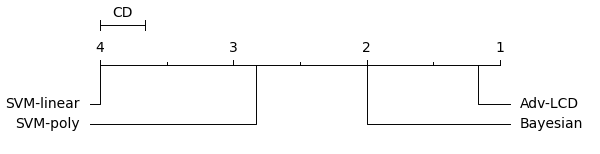}
\end{center}
  \caption{Critical Difference diagram }
  \label{fig:cd_diag}
\end{figure}

To assess the statistical robustness of our findings, we utilized Autorank \cite{autorank} to further analyze the performance of each algorithm. Autorank is an automated ranking algorithm that follows the guidelines proposed by Demšar \cite{cd_diagram} and employs independent paired samples to determine the differences in central tendency, such as median (MED), mean rank (MR), and median absolute deviation (MAD), for ranking each algorithm. It also provides the critical difference (CD), which is a statistical technique utilized to ascertain whether the performance difference between two or more algorithms is statistically significant. For the Autorank evaluation, we used an $\alpha = 0.05$. The result presented in Table \ref{tab:autorank} shows that \textit{Adversarial-LCD} received the highest ranking against the other ML algorithms, and Figure \ref{fig:cd_diag} shows that there is a critical difference between \textit{Adversarial-LCD} and the other algorithms, highlighting the effectiveness of our SVM-RBF surrogate model.

\begin{table} [h]
\tiny
\centering
\caption{The decline in LCD accuracy using different surrogate models}
\label{tab:surrogate}
\begin{tabular}{ccccc} \toprule \midrule
      
      Dataset &SVM-linear &SVM-Poly &Bayesian &Adv-LCD
      
      \\ \midrule
      fr2-desk          &-13.29  &-17.37  &-22.22 &\textbf{-27.92} 
      \\
      fr3-longoffice    &-14.10 &-19.72  &-26.94 &\textbf{-29.33}
      \\
      lounge            &-12.89  &-19.89 &-26.21 &\textbf{-28.72}
      \\
      kitchen           &-15.67  &-18.64 &\textbf{-24.99} &-24.66
      \\
      garden            &-13.22  &-17.89 &-18.46 &\textbf{-21.69} 
      \\ \midrule 
      Average           &-13.83  &-18.50 &-23.76 &\textbf{-26.46}
      \\
      \bottomrule
\end{tabular}
\end{table}
\begin{table}[h]
\tiny
\centering
\caption{Autorank analysis on different surrogate models}
\label{tab:autorank}
\begin{tabular}{lrrllll}
\toprule
{} &    MR &     MED &   MAD &                  CI & $\gamma$ &   Mag. \\
\midrule
\textbf{Adv-LCD}    & 3.83 & -27.19 & 1.83 &  [-27.9, -26.4] &     0.0 &  neg. \\
Bayesian   & 3.00 & -24.37 & 1.99 &  [-24.9, -23.7] &    -0.9 &       large \\
SVM-poly   & 2.16 & -19.41 & 0.48 &  [-19.7, -19.1] &    -3.9 &       large \\
SVM-linear & 1.00 & -13.56 & 0.44 &  [-13.8, -13.2] &    -6.8 &       large \\
\bottomrule
\end{tabular}
\end{table}

\subsection{Ablation study - the effectiveness of eigencentrality against other perturbation methods}
\label{sec:eigen}
\begin{figure} [t]
\begin{center}
\includegraphics[width=0.8\linewidth]{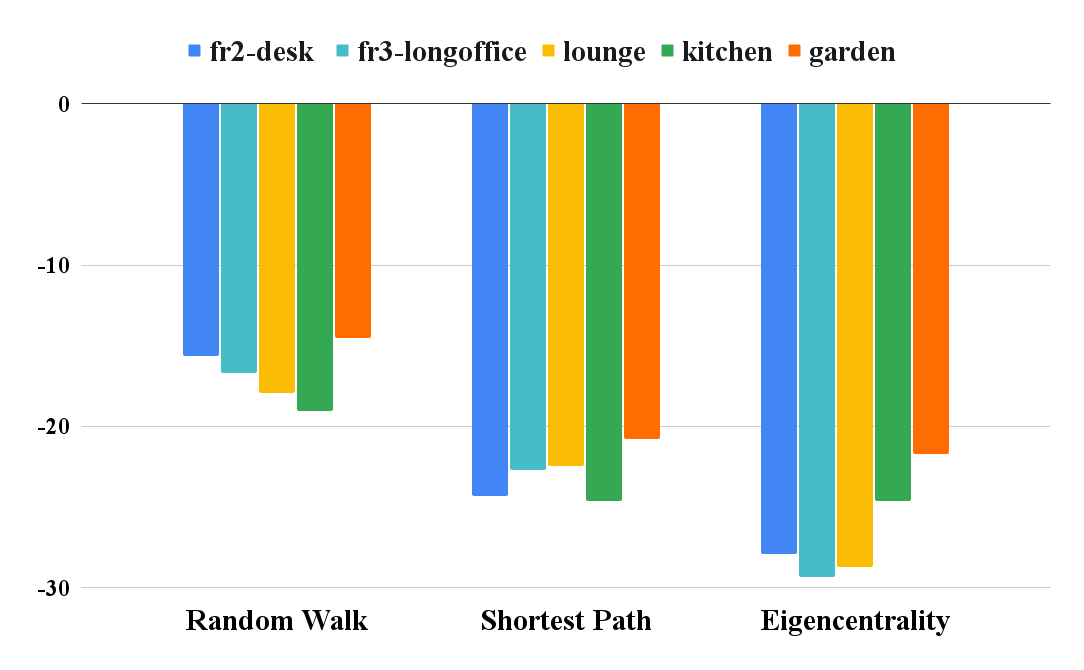}
\end{center}
   \caption{The decline in LCD accuracy using different perturbation methods}
   \label{fig:ablation_graph}
\end{figure}

We conducted two evaluations to assess the effectiveness of the eigencentrality perturbation method. To account for the non-deterministic nature of the algorithms, we performed both evaluations ten times. For the first evaluation, we kept all parameters, including the perturbation budget, identical to the previous evaluation. The results presented in Table \ref{tab:ablation} and Figure \ref{fig:ablation_graph} show that, on average, the eigencentrality method outperforms Random-walk by 9.6\% and Shortest-path by 4.0\%. 
The result demonstrates that our perturbation method based on node centrality is more effective than modifying shortest paths or selecting perturbations randomly.

For the second evaluation, we constrained the perturbation budgets further to compare the perturbation-efficiency of the eigencentrality method against other methods. The evaluation was performed using the \textit{fr2-desk} dataset.
The result presented in Table \ref{tab:pert_budget} shows that the eigencentrality method outperformed both Random-walk and Shortest-path across all evaluated perturbation budgets. On average, the eigencentrality method surpassed Random-walk by 7.9\% and Shortest-path by 3.1\%, highlighting the strong perturbation-efficiency of our method.

\begin{table} [h]
\tiny
\centering
\caption{The decline in LCD accuracy using different perturbation methods}
\label{tab:ablation}
\begin{tabular}{cccc} \toprule \midrule
     
      Dataset &Random Walk &Shortest Path & Eigencentrality
      
      \\ \midrule
      fr2-desk          &-15.66  &-24.30 &\textbf{-27.92} 
      \\
      fr3-longoffice    &-16.72 &-22.69  &\textbf{-29.33} 
      \\
      lounge            &-17.95  &-22.46 &\textbf{-28.72} 
      \\
      kitchen           &-19.04  &-24.63 &\textbf{-24.66}  
      \\
      garden            &-14.55  &-20.76 &\textbf{-21.69}
      \\ \midrule 
      Average           &-16.548  &-22.36 &\textbf{-26.46} 
      \\
      \bottomrule

\end{tabular}
\end{table}

\section{\uppercase{Conclusion and Future Work}}
\label{sec:conclusion}
In this paper, we presented \textit{Adversarial-LCD}, a novel \textit{black-box} \textit{evasion} attack framework, which uses an eigencentrality graph perturbation method and an SVM-RBF surrogate model with a Weisfeiler-Lehman feature extractor.
We showed that our \textit{Adversarial-LCD} with SVM-RBF surrogate model outperformed other ML surrogate algorithms, such as SVM-linear, SVM-polynomial and Bayesian classifier, demonstrating the effectiveness of \textit{Adversarial-LCD} framework. 

\begin{table} [h]
\tiny
\centering
\caption{The decline in LCD accuracy at different perturbation budgets (fr2-desk)}
\label{tab:pert_budget}
\begin{tabular}{cccc} \toprule \midrule
     
      Budget &Random Walk &Shortest Path & Eigencentrality
      
      \\ \midrule
       $r=1e^{-4}$         &-1.5  &-3.45 &\textbf{-5.02}
      \\
       $r=2e^{-4}$         &-7.33 &-11.22  &\textbf{-16.33} 
      \\
       $r=3e^{-4}$         &-15.66  &-24.30 &\textbf{-27.92} 
      \\ \midrule 
      Average           &-8.16  &-12.95 &\textbf{-16.09} 
      \\
      \bottomrule

\end{tabular}
\end{table}

Furthermore, we demonstrated that our perturbation method based on eigencentrality outperformed other algorithms such as Random-walk and Shortest-path in generating successful adversarial perturbations, highlighting that perturbing nodes based on their centrality is more efficient than randomly selecting perturbations or modifying the shortest paths between nodes.

Our future research will focus on exploring adversarial defence techniques, such as adversarial learning, for graph-based loop closure detection.

\section*{\uppercase{Acknowledgements}}
This research was supported by Callaghan Innovation, New Zealand government's Innovation Agency

\bibliographystyle{apalike}
{\small
\bibliography{example}}

\end{document}